# Universal Semantic Parsing

**Siva Reddy**[†*]  **Oscar Täckström**[‡]  **Slav Petrov**[‡]  **Mark Steedman**[††]  **Mirella Lapata**[††]

[†]Stanford University
[‡] Google Inc.
[††]University of Edinburgh

sivar@stanford.edu, {oscart, slav}@google.com, {steedman, mlap}@inf.ed.ac.uk

## Abstract

Universal Dependencies (UD) offer a uniform cross-lingual syntactic representation, with the aim of advancing multilingual applications. Recent work shows that semantic parsing can be accomplished by transforming syntactic dependencies to logical forms. However, this work is limited to English, and cannot process dependency graphs, which allow handling complex phenomena such as control. In this work, we introduce UDEPLAMBDA, a semantic interface for UD, which maps natural language to logical forms in an almost language-independent fashion and can process dependency graphs. We perform experiments on question answering against Freebase and provide German and Spanish translations of the WebQuestions and GraphQuestions datasets to facilitate multilingual evaluation. Results show that UDEPLAMBDA outperforms strong baselines across languages and datasets. For English, it achieves a 4.9 $F_1$ point improvement over the state-of-the-art on GraphQuestions.

## 1 Introduction

The Universal Dependencies (UD) initiative seeks to develop cross-linguistically consistent annotation guidelines as well as a large number of uniformly annotated treebanks for many languages (Nivre et al., 2016). Such resources could advance multilingual applications of parsing, improve comparability of evaluation results, enable cross-lingual learning, and more generally support natural language understanding.

Seeking to exploit the benefits of UD for natural language understanding, we introduce UDEPLAMBDA, a semantic interface for UD that maps natural language to logical forms, representing underlying predicate-argument structures, in an almost language-independent manner. Our framework is based on DEPLAMBDA (Reddy et al., 2016) a recently developed method that converts English Stanford Dependencies (SD) to logical forms. The conversion process is illustrated in Figure 1 and discussed in more detail in Section 2. Whereas DEPLAMBDA works only for English, U-DEPLAMBDA applies to any language for which UD annotations are available.[1] Moreover, DEPLAMBDA can only process tree-structured inputs whereas UDEPLAMBDA can also process dependency graphs, which allow to handle complex constructions such as control. The different treatments of various linguistic constructions in UD compared to SD also require different handling in UDEPLAMBDA (Section 3.3).

Our experiments focus on Freebase semantic parsing as a testbed for evaluating the framework's multilingual appeal. We convert natural language to logical forms which in turn are converted to machine interpretable formal meaning representations for retrieving answers to questions from Freebase. To facilitate multilingual evaluation, we provide translations of the English WebQuestions (Berant et al., 2013) and GraphQuestions (Su et al., 2016) datasets to German and Spanish. We demonstrate that UDEPLAMBDA can be used to derive logical forms for these languages using a minimal amount of language-specific knowledge. Aside from developing the first multilingual semantic parsing tool for Freebase, we also experimentally show that U-DEPLAMBDA outperforms strong baselines across languages and datasets. For English, it achieves the strongest result to date on GraphQuestions, with competitive results on WebQuestions. Our implementation and translated datasets are publicly available at https://github.com/sivareddyg/udeplambda.

---

[*]Work done at the University of Edinburgh

[1]As of v1.3, UD annotations are available for 47 languages at http://universaldependencies.org.

## 2 DEPLAMBDA

Before describing UDEPLAMBDA, we provide an overview of DEPLAMBDA (Reddy et al., 2016) on which our approach is based. DEPLAMBDA converts a dependency tree to its logical form in three steps: *binarization*, *substitution*, and *composition*, each of which is briefly outlined below. Algorithm 1 describes the steps of DEPLAMBDA in lines 4-6, whereas lines 2 and 3 are specific to UDEPLAMBDA.

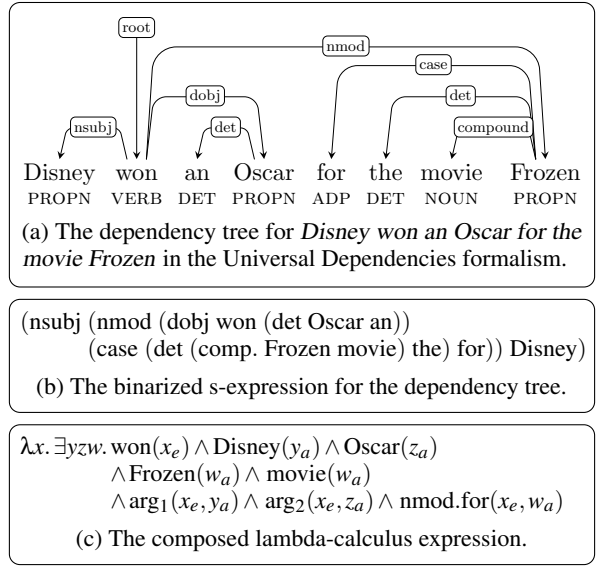

(a) The dependency tree for *Disney won an Oscar for the movie Frozen* in the Universal Dependencies formalism.

(nsubj (nmod (dobj won (det Oscar an))
    (case (det (comp. Frozen movie) the) for)) Disney)

(b) The binarized s-expression for the dependency tree.

$\lambda x. \exists yzw. \text{won}(x_e) \wedge \text{Disney}(y_a) \wedge \text{Oscar}(z_a)$
$\wedge \text{Frozen}(w_a) \wedge \text{movie}(w_a)$
$\wedge \arg_1(x_e, y_a) \wedge \arg_2(x_e, z_a) \wedge \text{nmod.for}(x_e, w_a)$

(c) The composed lambda-calculus expression.

Figure 1: The mapping of a dependency tree to its logical form with the intermediate s-expression.

**Binarization** A dependency tree is first mapped to a Lisp-style s-expression indicating the order of semantic composition. Figure 1(b) shows the s-expression for the sentence *Disney won an Oscar for the movie Frozen*, derived from the dependency tree in Figure 1(a). Here, the sub-expression (dobj won (det Oscar an)) indicates that the logical form of the phrase *won an Oscar* is derived by composing the logical form of the label dobj with the logical form of the word *won* and the logical form of the phrase *an Oscar*, derived analogously. The s-expression can also be interpreted as a binarized tree with the dependency label as the root node, and the left and right expressions as subtrees.

A composition hierarchy is employed to impose a strict traversal ordering on the modifiers to each head in the dependency tree. As an example, *won* has three modifiers in Figure 1(a), which according to the composition hierarchy are composed in the order dobj > nmod > nsubj. In constructions like coordination, this ordering is crucial to arrive at the correct semantics. Lines 7-17 in Algorithm 1 describe the binarization step.

**Substitution** Each symbol in the s-expressions is substituted for a lambda expression encoding its semantics. Words and dependency labels are assigned different types of expressions. In general, words have expressions of the following kind:

ENTITY $\Rightarrow \lambda x. \text{word}(x_a)$; e.g. Oscar $\Rightarrow \lambda x. \text{Oscar}(x_a)$
EVENT $\Rightarrow \lambda x. \text{word}(x_e)$; e.g. won $\Rightarrow \lambda x. \text{won}(x_e)$
FUNCTIONAL $\Rightarrow \lambda x. \text{TRUE}$; e.g. an $\Rightarrow \lambda x. \text{TRUE}$

Here, the subscripts $\cdot_a$ and $\cdot_e$ denote the types of individuals (**Ind**) and events (**Event**), respectively, whereas $x$ denotes a paired variable $(x_a, x_e)$ of type **Ind** × **Event**. Roughly speaking, proper nouns and adjectives invoke ENTITY expressions, verbs and adverbs invoke EVENT expressions, and common nouns invoke both ENTITY and EVENT expressions (see Section 3.3), while remaining words invoke FUNCTIONAL expressions. DEPLAMBDA enforces the constraint that every s-expression is of the type $\eta = \text{Ind} \times \text{Event} \rightarrow \text{Bool}$, which simplifies the type system considerably.

Expressions for dependency labels glue the semantics of heads and modifiers to articulate predicate-argument structure. These expressions in general take one of the following forms:

COPY $\Rightarrow \lambda fgx. \exists y. f(x) \wedge g(y) \wedge \text{rel}(x, y)$
e.g. nsubj, dobj, nmod, advmod
INVERT $\Rightarrow \lambda fgx. \exists y. f(x) \wedge g(y) \wedge \text{rel}^i(y, x)$
e.g. amod, acl
MERGE $\Rightarrow \lambda fgx. f(x) \wedge g(x)$
e.g. compound, appos, amod, acl
HEAD $\Rightarrow \lambda fgx. f(x)$
e.g. case, punct, aux, mark.

As an example of COPY, consider the lambda expression for dobj in (dobj won (det Oscar an)): $\lambda fgx. \exists y. f(x) \wedge g(y) \wedge \arg_2(x_e, y_a)$. This expression takes two functions $f$ and $g$ as input, where $f$ represents the logical form of *won* and $g$ represents the logical form of *an Oscar*. The predicate-argument structure $\arg_2(x_e, y_a)$ indicates that the $\arg_2$ of the event $x_e$, i.e. *won*, is the individual $y_a$, i.e. the entity *Oscar*. Since $\arg_2(x_e, y_a)$ mimics the dependency structure dobj(won, Oscar), we refer to the expression kind evoked by dobj as COPY.

Expressions that invert the dependency direction are referred to as INVERT (e.g. amod in *running horse*); expressions that merge two subexpressions without introducing any relation predicates are referred to as MERGE (e.g. compound in *movie Frozen*); and expressions that simply return the parent expression semantics are referred to as HEAD (e.g. case in *for Frozen*). While this generalization applies to most dependency labels, several labels take a different logical form not listed here, some of which are discussed in Section 3.3. Sometimes the mapping of dependency label to lambda expression may depend on surrounding part-of-speech tags or dependency labels. For example, amod acts as INVERT when the modifier is a verb (e.g. in *running horse*), and as MERGE when the modifier is an adjective (e.g. in *beautiful horse*).[2] Lines 26-32 in Algorithm 1 describe the substitution procedure.

**Composition** The final logical form is computed by beta-reduction, treating expressions of the form (f x y) as the function f applied to the arguments x and y. For example, (dobj won (det Oscar an)) results in $\lambda x. \exists z. \text{won}(x_e) \wedge \text{Oscar}(z_a) \wedge \arg_2(x_e, z_a)$ when the expression for dobj is applied to those for *won* and *(det Oscar an)*. Figure 1(c) shows the logical form for the s-expression in Figure 1(b). The binarized s-expression is recursively converted to a logical form as described in lines 18-25 in Algorithm 1.

## 3 UDEPLAMBDA

We now introduce UDEPLAMBDA, a semantic interface for Universal Dependencies.[3] Whereas DEPLAMBDA only applies to English Stanford Dependencies, UDEPLAMBDA takes advantage of the cross-lingual nature of UD to facilitate an (almost) language independent semantic interface. This is accomplished by restricting the binarization, substitution, and composition steps described above to rely solely on information encoded in the UD representation. As shown in Algorithm 1, lines 4-6 are common to both DEPLAMBDA and UDEPLAMBDA, whereas lines 2 and 3 applies only to UDEPLAMBDA. Importantly, UDEPLAMBDA is designed to not rely on lexical forms in a language

---

[2] We use Tregex (Levy and Andrew, 2006) for substitution mappings and Cornell SPF (Artzi, 2013) as the lambda-calculus implementation. For example, in *running horse*, the tregex */label:amod/=target < /postag:verb/* matches amod to its INVERT expression $\lambda fgx. \exists y. f(x) \wedge g(y) \wedge \text{amod}^i(y_e, x_a)$.

[3] In what follows, all references to UD are to UD v1.3.

---

**Algorithm 1:** UDEPLAMBDA Steps

1 **Function** UDepLambda (*depTree*):
2     *depGraph* = Enhancement (*depTree*)
     #See Figure 2(a) for a depGraph.
3     *bindedTree* = SplitLongDistance (*depGraph*)
     #See Figure 2(b) for a bindedTree.
4     *binarizedTree* = Binarization (*bindedTree*)
     #See Figure 1(b) for a binarizedTree.
5     *logicalForm* = Composition (*binarizedTree*)
6     **return** *logicalForm*

7 **Function** Binarization (*tree*):
8     *parent* = GetRootNode (*tree*);
9     $\{(label1, child1), (label2, child2) \ldots\}$
      = GetChildNodes (*parent*)
10     *sortedChildren* = SortUsingLabelHierarchy
      ($\{(label1, child1), (label2, child2) \ldots\}$)
11     *binarziedTree.root* = *parent*
12     **for** *label*, *child* $\in$ *sortedChildren*:
13        *temp.root* = *label*
14        *temp.left* = *binarziedTree*
15        *temp.right* = Binarization(*child*)
16        *binarziedTree* = *temp*
17     **return** *binarziedTree*

18 **Function** Composition (*binarizedTree*):
19     *mainLF* = Substitution (*binarizedTree.root*)
20     **if** *binarziedTree has left and right children*:
21        *leftLF* = Composition (*binarziedTree.left*)
22        *rightLF* = Composition(*binarziedTree.right*)
23        *mainLF* = BetaReduce (*mainLF*, *leftLF*)
24        *mainLF* = BetaReduce (*mainLF*, *rightLF*)
25     **return** *mainLF*

26 **Function** Substitution (*node*):
27     *logicalForms* = [ ]
28     **for** *tregexRule*, *template* $\in$ *substitutionRules*:
29        **if** *tregexRule.match(node)*:
30           *lf* = GenLambdaExp (node, template)
31           *logicalForms.add(lf)*
32     **return** *logicalForms*

---

to assign lambda expressions, but only on information contained in dependency labels and postags.

However, some linguistic phenomena are language specific (e.g. pronoun-dropping) or lexicalized (e.g. *every* and *the* in English have different semantics, despite being both determiners) and are not encoded in the UD schema. Furthermore, some cross-linguistic phenomena, such as long-distance dependencies, are not part of the core UD representation. To circumvent this limitation, a simple *enhancement* step enriches the original UD representation before binarization takes place (Section 3.1). This step adds to the dependency tree missing syntactic information and long-distance dependencies, thereby creating a graph. Whereas DEPLAMBDA is not able to handle graph-structured input, UDEP-

LAMBDA is designed to work with dependency graphs as well (Section 3.2). Finally, several constructions differ in structure between UD and SD, which requires different handling in the semantic interface (Section 3.3).

## 3.1 Enhancement

Both Schuster and Manning (2016) and Nivre et al. (2016) note the necessity of an enhanced UD representation to enable semantic applications. However, such enhancements are currently only available for a subset of languages in UD. Instead, we rely on a small number of enhancements for our main application—semantic parsing for question-answering—with the hope that this step can be replaced by an enhanced UD representation in the future. Specifically, we define three kinds of enhancements: (1) long-distance dependencies; (2) types of coordination; and (3) refined question word tags. These correspond to line 2 in Algorithm 1.

First, we identify long-distance dependencies in relative clauses and control constructions. We follow Schuster and Manning (2016) and find these using the labels acl (relative) and xcomp (control). Figure 2(a) shows the long-distance dependency in the sentence *Anna wants to marry Kristoff*. Here, *marry* is provided with its missing nsubj (dashed arc). Second, UD conflates all coordinating constructions to a single dependency label, conj. To obtain the correct coordination scope, we refine conj to conj:verb, conj:vp, conj:sentence, conj:np, and conj:adj, similar to Reddy et al. (2016). Finally, unlike the PTB tags (Marcus et al., 1993) used by SD, the UD part-of-speech tags do not distinguish question words. Since these are crucial to question-answering, we use a small lexicon to refine the tags for determiners (DET), adverbs (ADV) and pronouns (PRON) to DET:WH, ADV:WH and PRON:WH, respectively. Specifically, we use a list of 12 (English), 14 (Spanish) and 35 (German) words, respectively. This is the only part of UDEPLAMBDA that relies on language-specific information. We hope that, as the coverage of morphological features in UD improves, this refinement can be replaced by relying on morphological features, such as the interrogative feature (INT).

## 3.2 Graph Structures and BIND

To handle graph structures that may result from the enhancement step, such as those in Figure 2(a), we propose a variable-binding mechanism that differs

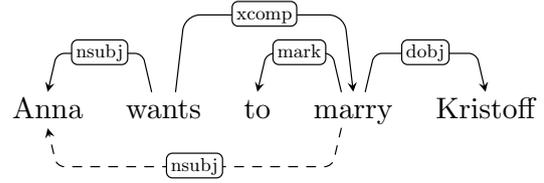

(a) With long-distance dependency.

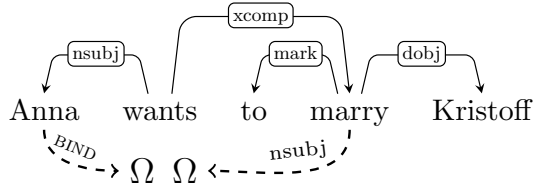

(b) With variable binding.

Figure 2: The original and enhanced dependency trees for *Anna wants to marry Kristoff*.

from that of DEPLAMBDA. This is indicated in line 3 of Algorithm 1. First, each long-distance dependency is split into independent arcs as shown in Figure 2(b). Here, $\Omega$ is a placeholder for the subject of *marry*, which in turn corresponds to *Anna* as indicated by the binding of $\Omega$ via the pseudo-label BIND. We treat BIND like an ordinary dependency label with semantics MERGE and process the resulting tree as usual, via the s-expression:

(nsubj (xcomp wants (nsubj (mark
  (dobj marry Kristoff) to) $\Omega$) (BIND Anna $\Omega$)),

with the lambda-expression substitutions:

*wants*, *marry* ∈ EVENT; *to* ∈ FUNCTIONAL;
*Anna*, *Kristoff* ∈ ENTITY;
mark ∈ HEAD; BIND ∈ MERGE;
xcomp = $\lambda f g x. \exists y. f(x) \wedge g(y) \wedge \text{xcomp}(x_e, y_e)$.

These substitutions are based solely on unlexicalized context. For example, the part-of-speech tag PROPN of *Anna* invokes an ENTITY expression.

The placeholder $\Omega$ has semantics $\lambda x.\text{EQ}(x,\omega)$, where $\text{EQ}(u,\omega)$ is true iff $u$ and $\omega$ are equal (have the same denotation), which unifies the subject variable of *wants* with the subject variable of *marry*.

After substitution and composition, we get:

$\lambda z. \exists xywv. \text{wants}(z_e) \wedge \text{Anna}(x_a) \wedge \text{arg}_1(z_e, x_a) \wedge \text{EQ}(x,\omega)$
$\wedge \text{marry}(y_e) \wedge \text{xcomp}(z_e, y_e) \wedge \text{arg}_1(y_e, v_a) \wedge \text{EQ}(v,\omega)$
$\wedge \text{Kristoff}(w_a) \wedge \text{arg}_2(y_e, w_a)$,

This expression may be simplified further by replacing all occurrences of $v$ with $x$ and removing the unification predicates EQ, which results in:

$\lambda z. \exists xyw. \text{wants}(z_e) \wedge \text{Anna}(x_a) \wedge \text{arg}_1(z_e, x_a)$
$\wedge \text{marry}(y_e) \wedge \text{xcomp}(z_e, y_e) \wedge \text{arg}_1(y_e, x_a)$
$\wedge \text{Kristoff}(w_a) \wedge \text{arg}_2(y_e, w_a)$.

This expression encodes the fact that *Anna* is the arg$_1$ of the *marry* event, as desired. DEPLAMBDA, in contrast, cannot handle graph-structured input, since it lacks a principled way of generating s-expressions from graphs. Even given the above s-expression, BIND in DEPLAMBDA is defined in a way such that the composition fails to unify *v* and *x*, which is crucial for the correct semantics. Moreover, the definition of BIND in DEPLAMBDA does not have a formal interpretation within the lambda calculus, unlike ours.

### 3.3 Linguistic Constructions

Below, we highlight the most pertinent differences between UDEPLAMBDA and DEPLAMBDA, stemming from the different treatment of various linguistic constructions in UD versus SD.

**Prepositional Phrases** UD uses a content-head analysis, in contrast to SD, which treats function words as heads of prepositional phrases, Accordingly, the s-expression for the phrase *president in 2009* is (nmod president (case 2009 in)) in U-DEPLAMBDA and (prep president (pobj in 2009)) in DEPLAMBDA. To achieve the desired semantics,

$\lambda x. \exists y.\, \text{president}(x_a) \wedge \text{president\_event}(x_e) \wedge \arg_1(x_e, x_a) \wedge 2009(y_a) \wedge \text{prep.in}(x_e, y_a)\,,$

DEPLAMBDA relies on an intermediate logical form that requires some post-processing, whereas UDEPLAMBDA obtains the desired logical form directly through the following entries:

*in* ∈ FUNCTIONAL; *2009* ∈ ENTITY; *case* ∈ HEAD;
*president* = $\lambda x.\, \text{president}(x_a) \wedge \text{president\_event}(x_e)$
$\qquad\qquad \wedge \arg_1(x_e, x_a)$ ;
*nmod* = $\lambda fgx.\, \exists y.\, f(x) \wedge g(y) \wedge \text{nmod.in}(x_e, y_a)$ .

Other `nmod` constructions, such as possessives (`nmod:poss`), temporal modifiers (`nmod:tmod`) and adverbial modifiers (`nmod:npmod`), are handled similarly. Note how the common noun *president*, evokes both entity and event predicates above.

**Passives** DEPLAMBDA gives special treatment to passive verbs, identified by the fine-grained part-of-speech tags in the PTB tag together with dependency context. For example, *An Oscar was won* is analyzed as $\lambda x.\, \text{won.pass}(x_e) \wedge \text{Oscar}(y_a) \wedge \arg_1(x_e, y_a)$, where won.pass represents a passive event. However, UD does not distinguish between active and passive forms.[4] While the labels `nsubjpass` or `auxpass` indicate passive constructions, such clues are sometimes missing, such as in reduced relatives. We therefore opt to not have separate entries for passives, but aim to produce identical logical forms for active and passive forms when possible (for example, by treating `nsubjpass` as direct object). With the following entries,

*won* ∈ EVENT; *an*, *was* ∈ FUNCTIONAL; *auxpass* ∈ HEAD;
*nsubjpass* = $\lambda fgx.\, \exists y.\, f(x) \wedge g(y) \wedge \arg_2(x_e, y_a)$,

the lambda expression for *An Oscar was won* becomes $\lambda x.\, \text{won}(x_e) \wedge \text{Oscar}(y_a) \wedge \arg_2(x_e, y_a)$, identical to that of its active form. However, not having a special entry for passive verbs may have undesirable side-effects. For example, in the reduced-relative construction *Pixar claimed the Oscar won for Frozen*, the phrase *the Oscar won ...* will receive the semantics $\lambda x.\, \text{Oscar}(y_a) \wedge \text{won}(x_e) \wedge \mathbf{arg_1}(x_e, y_a)$, which differs from that of *an Oscar was won*. We leave it to the target application to disambiguate the interpretation in such cases.

**Long-Distance Dependencies** As discussed in Section 3.2, we handle long-distance dependencies evoked by clausal modifiers (`acl`) and control verbs (`xcomp`) with the BIND mechanism, whereas DEPLAMBDA cannot handle control constructions. For `xcomp`, as seen earlier, we use the mapping $\lambda fgx.\, \exists y.\, f(x) \wedge g(y) \wedge \text{xcomp}(x_e, y_e)$. For `acl` we use $\lambda fgx.\, \exists y.\, f(x) \wedge g(y)$, to conjoin the main clause and the modifier clause. However, not all `acl` clauses evoke long-distance dependencies, e.g. in *the news that Disney won an Oscar*, the clause *that Disney won an Oscar* is a subordinating conjunction of *news*. In such cases, we instead assign `acl` the INVERT semantics.

**Questions** Question words are marked with the enhanced part-of-speech tags DET:WH, ADV:WH and PRON:WH, which are all assigned the semantics $\lambda x.\, \${\text{word}\}(x_a) \wedge \text{TARGET}(x_a)$. The predicate TARGET indicates that $x_a$ represents the variable of interest, that is the answer to the question.

### 3.4 Limitations

In order to achieve language independence, UDEPLAMBDA has to sacrifice semantic specificity, since in many cases the semantics is carried by lexical information. Consider the sentences *John broke the window* and *The window broke*. Although it is the *window* that broke in both cases, our inferred logical forms do not canonicalize the relation between *broke* and *window*. To achieve this, we would have to make the substitution of nsubj depend on lexical context, such that when *window* occurs as nsubj with *broke*, the predicate arg$_2$ is invoked rather than arg$_1$. UDEPLAMBDA does not address this problem, and leave it to the target application to infer context-sensitive semantics of arg$_1$ and arg$_2$. To measure the impact of this limitation, we present UDEPLAMBDASRL in Section 4.4 which addresses this problem by relying on semantic roles from semantic role labeling (Palmer et al., 2010).

---

[4] UD encodes voice as a morphological feature, but most syntactic analyzers do not produce this information yet.

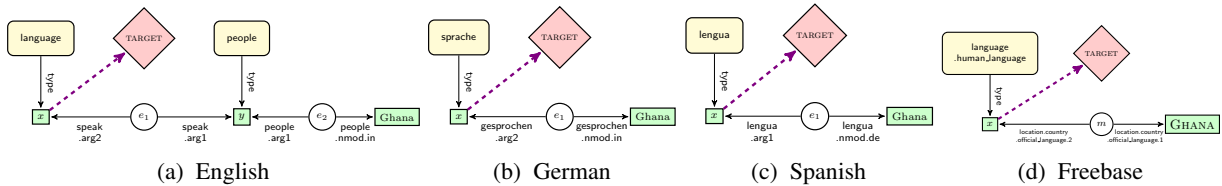

Figure 3: The ungrounded graphs for *What language do the people in Ghana speak?*, *Welche Sprache wird in Ghana gesprochen?* and *Cuál es la lengua de Ghana?*, and the corresponding grounded graph.

Other constructions that require lexical information are quantifiers like *every*, *some* and *most*, negation markers like *no* and *not*, and intentional verbs like *believe* and *said*. UD does not have special labels to indicate these. We discuss how to handle quantifiers in this framework in the supplementary material.

Although in the current setup UDEPLAMBDA rules are hand-coded, the number of rules are only proportional to the number of UD labels, making rule-writing manageable.[5] Moreover, we view UDEPLAMBDA as a first step towards learning rules for converting UD to richer semantic representations such as PropBank, AMR, or the Parallel Meaning Bank (Palmer et al., 2005; Banarescu et al., 2013; Abzianidze et al., 2017)..

## 4 Cross-lingual Semantic Parsing

To study the multilingual nature of UDEPLAMBDA, we conduct an empirical evaluation on question answering against Freebase in three different languages: English, Spanish, and German. Before discussing the details of this experiment, we briefly outline the semantic parsing framework employed.

---

[5] UD v1.3 has 40 dependency labels, and the number of substitution rules in UDEPLAMBDA are 61, with some labels having multiple rules, and some representing lexical semantics.

### 4.1 Semantic Parsing as Graph Matching

UDEPLAMBDA generates *ungrounded* logical forms that are independent of any knowledge base, such as Freebase. We use GRAPHPARSER (Reddy et al., 2016) to map these logical forms to their grounded Freebase graphs, via corresponding ungrounded graphs. Figures 3(a) to 3(c) show the ungrounded graphs corresponding to logical forms from UDEPLAMBDA, each grounded to the same Freebase graph in Figure 3(d). Here, rectangles denote entities, circles denote events, rounded rectangles denote entity types, and edges between events and entities denote predicates or Freebase relations. Finally, the TARGET node represents the set of values of *x* that are consistent with the Freebase graph, that is the answer to the question.

GRAPHPARSER treats semantic parsing as a graph-matching problem with the goal of finding the Freebase graphs that are structurally isomorphic to an ungrounded graph and rank them according to a model. To account for structural mismatches, GRAPHPARSER uses two graph transformations: CONTRACT and EXPAND. In Figure 3(a) there are two edges between *x* and *Ghana*. CONTRACT collapses one of these edges to create a graph isomorphic to Freebase. EXPAND, in contrast, adds edges to connect the graph in the case of disconnected components. The search space is explored by beam search and model parameters are estimated with the averaged structured perceptron (Collins, 2002) from training data consisting of question-answer pairs, using answer $F_1$-score as the objective.

### 4.2 Datasets

We evaluate our approach on two public benchmarks of question answering against Freebase: WebQuestions (Berant et al., 2013), a widely used benchmark consisting of English questions and their answers, and GraphQuestions (Su et al., 2016), a recently released dataset of English questions with both their answers and grounded logical forms.

While WebQuestions is dominated by simple entity-attribute questions, GraphQuestions contains a large number of compositional questions involving aggregation (e.g. *How many children of Eddard Stark were born in Winterfell?*) and comparison (e.g. *In which month does the average rainfall of New York City exceed 86 mm?*). The number of training, development and test questions is 2644, 1134, and 2032, respectively, for WebQuestions and 1794, 764, and 2608 for GraphQuestions.

To support multilingual evaluation, we created translations of WebQuestions and GraphQuestions to German and Spanish. For WebQuestions two professional annotators were hired per language, while for GraphQuestions we used a trusted pool of 20 annotators per language (with a single annotator per question). Examples of the original questions and their translations are provided in Table 1.

### 4.3 Implementation Details

Here we provide details on the syntactic analyzers employed, our entity resolution algorithm, and the features used by the grounding model.

**Dependency Parsing** The English, Spanish, and German Universal Dependencies (UD) treebanks (v1.3; Nivre et al 2016) were used to train part of speech taggers and dependency parsers. We used a bidirectional LSTM tagger (Plank et al., 2016) and a bidirectional LSTM shift-reduce parser (Kiperwasser and Goldberg, 2016). Both the tagger and parser require word embeddings. For English, we used GloVe embeddings (Pennington et al., 2014) trained on Wikipedia and the Gigaword corpus. For German and Spanish, we used SENNA embeddings (Collobert et al., 2011; Al-Rfou et al., 2013) trained on Wikipedia corpora (589M words German; 397M words Spanish).[6] Measured on the UD test sets, the tagger accuracies are 94.5 (English), 92.2 (German), and 95.7 (Spanish), with corresponding labeled attachment parser scores of 81.8, 74.7, and 82.2.

**Entity Resolution** We follow Reddy et al. (2016) and resolve entities in three steps: (1) potential entity spans are identified using seven handcrafted part-of-speech patterns; (2) each span is associated with potential Freebase entities according to the Freebase/KG API; and (3) the 10-best entity linking lattices, scored by a structured perceptron, are input to GRAPHPARSER, leaving the final disambiguation to the semantic parsing problem. Table 2 shows the 1-best and 10-best entity disambiguation $F_1$-scores for each language and dataset.

|     | WebQuestions |
| --- | --- |
| en | What language do the people in Ghana speak? |
| de | Welche Sprache wird in Ghana gesprochen? |
| es | ¿Cuál es la lengua de Ghana? |
| en | Who was Vincent van Gogh inspired by? |
| de | Von wem wurde Vincent van Gogh inspiriert? |
| es | ¿Qué inspiró a Van Gogh? |
|     | GraphQuestions |
| en | NASA has how many launch sites? |
| de | Wie viele Abschussbasen besitzt NASA? |
| es | ¿Cuántos sitios de despegue tiene NASA? |
| en | Which loudspeakers are heavier than 82.0 kg? |
| de | Welche Lautsprecher sind schwerer als 82.0 kg? |
| es | ¿Qué altavoces pesan más de 82.0 kg? |

Table 1: Example questions and their translations.

| k | WebQuestions | | | GraphQuestions | | |
| --- | --- | --- | --- | --- | --- | --- |
|   | en | de | es | en | de | es |
| 1 | 89.6 | 82.8 | 86.7 | 47.2 | 39.9 | 39.5 |
| 10 | 95.7 | 91.2 | 94.0 | 56.9 | 48.4 | 51.6 |

Table 2: Structured perceptron *k*-best entity linking accuracies on the development sets.

**Features** We use features similar to Reddy et al. (2016): *basic* features of words and Freebase relations, and *graph* features crossing ungrounded events with grounded relations, ungrounded types with grounded relations, and ungrounded answer type crossed with a binary feature indicating if the answer is a number. In addition, we add features encoding the *semantic* similarity of ungrounded events and Freebase relations. Specifically, we used the cosine similarity of the translation-invariant embeddings of Huang et al. (2015).[7]

### 4.4 Comparison Systems

We compared UDEPLAMBDA to four versions of GRAPHPARSER that operate on different representations, in addition to prior work.

**SINGLEEVENT** This model resembles the learning-to-rank model of Bast and Haussmann (2015). An ungrounded graph is generated by connecting all entities in the question with the TARGET node, representing a single event. Note that this baseline cannot handle compositional questions, or those with aggregation or comparison.

---
[6] https://sites.google.com/site/rmyeid/projects/polyglot.

[7] http://128.2.220.95/multilingual/data/.

| Method | WebQuestions en | de | es | GraphQuestions en | de | es |
|---|---|---|---|---|---|---|
| SINGLEEVENT | 48.5 | 45.6 | 46.3 | 15.9 | 8.8 | 11.4 |
| DEPTREE | 48.8 | 45.9 | 46.4 | 16.0 | 8.3 | 11.3 |
| CCGGRAPH | 49.5 | – | – | 15.9 | – | – |
| UDEPLAMBDA | 49.5 | 46.1 | 47.5 | 17.7 | 9.5 | 12.8 |
| UDEPLAMBDASRL | 49.8 | 46.2 | 47.0 | 17.7 | 9.1 | 12.7 |

Table 3: $F_1$-scores on the test data.

**DEPTREE** An ungrounded graph is obtained directly from the original dependency tree. An event is created for each parent and its dependents in the tree. Each dependent is linked to this event with an edge labeled with its dependency relation, while the parent is linked to the event with an edge labeled $arg_0$. If a word is a question word, an additional TARGET predicate is attached to its entity node.

**CCGGRAPH** This is the CCG-based semantic representation of Reddy et al. (2014). Note that this baseline exists only for English.

**UDEPLAMBDASRL** This is similar to UDEPLAMBDA except that instead of assuming nsubj, dobj and nsubjpass correspond to $arg_1$, $arg_2$ and $arg_2$, we employ semantic role labeling to identify the correct interpretation. We used the systems of Roth and Woodsend (2014) for English and German and Bjrkelund et al. (2009) for Spanish trained on the CoNLL-2009 dataset (Haji et al., 2009).[8]

### 4.5 Results

Table 3 shows the performance of GRAPHPARSER with these different representations. Here and in what follows, we use average $F_1$-score of predicted answers (Berant et al., 2013) as the evaluation metric. We first observe that UDEPLAMBDA consistently outperforms the SINGLEEVENT and DEPTREE representations in all languages.

For English, performance is on par with CCG-GRAPH, which suggests that UDEPLAMBDA does not sacrifice too much specificity for universality. With both datasets, results are lower for German compared to Spanish. This agrees with the lower performance of the syntactic parser on the German portion of the UD treebank. While U-DEPLAMBDASRL performs better than UDEP-

---

[8] The parser accuracies (%) are 87.33, 81.38 and 79.91 for English, German and Spanish respectively.

| Method | GraphQ. | WebQ. |
|---|---|---|
| SEMPRE (Berant et al., 2013) | 10.8 | 35.7 |
| JACANA (Yao and Van Durme, 2014) | 5.1 | 33.0 |
| PARASEMPRE (Berant and Liang, 2014) | 12.8 | 39.9 |
| QA (Yao, 2015) | – | 44.3 |
| AQQU (Bast and Haussmann, 2015) | – | 49.4 |
| AGENDAIL (Berant and Liang, 2015) | – | 49.7 |
| DEPLAMBDA (Reddy et al., 2016) | – | 50.3 |
| STAGG (Yih et al., 2015) | – | 48.4 (52.5) |
| BILSTM (Türe and Jojic, 2016) | – | 24.9 (52.2) |
| MCNN (Xu et al., 2016) | – | 47.0 (53.3) |
| AGENDAIL-RANK (Yavuz et al., 2016) | – | 51.6 (52.6) |
| UDEPLAMBDA | 17.7 | 49.5 |

Table 4: $F_1$-scores on the English GraphQuestions and WebQuestions test sets (results with additional task-specific resources in parentheses).

LAMBDA on WebQuestions for English, we do not see large performance gaps in other settings, suggesting that GRAPHPARSER is either able to learn context-sensitive semantics of ungrounded predicates or that the datasets do not contain ambiguous nsubj, dobj and nsubjpass mappings. Finally, while these results confirm that GraphQuestions is much harder compared to WebQuestions, we note that both datasets predominantly contain single-hop questions, as indicated by the competitive performance of SINGLEEVENT on both datasets.

Table 4 compares UDEPLAMBDA with previously published models which exist only for English and have been mainly evaluated on WebQuestions. These are either symbolic like ours (first block) or employ neural networks (second block). Results for models using additional task-specific training resources, such as ClueWeb09, Wikipedia, or SimpleQuestions (Bordes et al., 2015) are shown in parentheses. On GraphQuestions, we achieve a new state-of-the-art result with a gain of 4.8 $F_1$-points over the previously reported best result. On WebQuestions we are 2.1 points below the best model using comparable resources, and 3.8 points below the state of the art. Most related to our work is the English-specific system of Reddy et al. (2016). We attribute the 0.8 point difference in $F_1$-score to their use of the more fine-grained PTB tag set and Stanford Dependencies.

## 5 Related Work

Our work continues the long tradition of building logical forms from syntactic representations initiated by Montague (1973). The literature is rife with attempts to develop semantic interfaces for HPSG (Copestake et al., 2005), LFG (Kaplan and Bresnan, 1982; Dalrymple et al., 1995; Crouch and King, 2006), TAG (Kallmeyer and Joshi, 2003; Gardent and Kallmeyer, 2003; Nesson and Shieber, 2006), and CCG (Baldridge and Kruijff, 2002; Bos et al., 2004; Artzi et al., 2015). Unlike existing semantic interfaces, UDEPLAMBDA uses dependency syntax, a widely available syntactic resource.

A common trend in previous work on semantic interfaces is the reliance on rich typed feature structures or semantic types coupled with strong type constraints, which can be very informative but unavoidably language specific. Instead, UDEPLAMBDA relies on generic unlexicalized information present in dependency treebanks and uses a simple type system (one type for dependency labels, and one for words) along with a combinatory mechanism, which avoids type collisions. Earlier attempts at extracting semantic representations from dependencies have mainly focused on language-specific dependency representations (Spreyer and Frank, 2005; Simov and Osenova, 2011; Hahn and Meurers, 2011; Reddy et al., 2016; Falke et al., 2016; Beltagy, 2016), and multi-layered dependency annotations (Jakob et al., 2010; Bédaride and Gardent, 2011). In contrast, UDEPLAMBDA derives semantic representations for multiple languages in a common schema directly from Universal Dependencies. This work parallels a growing interest in creating other forms of multilingual semantic representations (Akbik et al., 2015; Vanderwende et al., 2015; White et al., 2016; Evang and Bos, 2016).

We evaluate UDEPLAMBDA on semantic parsing for question answering against a knowledge base. Here, the literature offers two main modeling paradigms: (1) learning of task-specific grammars that directly parse language to a grounded representation (Zelle and Mooney, 1996; Zettlemoyer and Collins, 2005; Berant et al., 2013; Flanigan et al., 2014; Pasupat and Liang, 2015; Groschwitz et al., 2015); and (2) converting language to a linguistically motivated task-independent representation that is then mapped to a grounded representation (Kwiatkowski et al., 2013; Reddy et al., 2014; Krishnamurthy and Mitchell, 2015; Gardner and Krishnamurthy, 2017). Our work belongs to the latter paradigm, as we map natural language to Freebase indirectly via logical forms. Capitalizing on natural-language syntax affords interpretability, scalability, and reduced duplication of effort across applications (Bender et al., 2015). Our work also relates to literature on parsing multiple languages to a common executable representation (Cimiano et al., 2013; Haas and Riezler, 2016). However, existing approaches still map to the target meaning representations (more or less) directly (Kwiatkowksi et al., 2010; Jones et al., 2012; Jie and Lu, 2014).

## 6 Conclusions

We introduced UDEPLAMBDA, a semantic interface for Universal Dependencies, and showed that the resulting semantic representation can be used for question-answering against a knowledge base in multiple languages. We provided translations of benchmark datasets in German and Spanish, in the hope to stimulate further multilingual research on semantic parsing and question answering in general. We have only scratched the surface when it comes to applying UDEPLAMBDA to natural language understanding tasks. In the future, we would like to explore how this framework can benefit applications such as summarization (Liu et al., 2015) and machine reading (Sachan and Xing, 2016).

**Acknowledgements**

This work greatly benefited from discussions with Michael Collins, Dipanjan Das, Federico Fancellu, Julia Hockenmaier, Tom Kwiatkowski, Adam Lopez, Valeria de Paiva, Martha Palmer, Fernando Pereira, Emily Pitler, Vijay Saraswat, Nathan Schneider, Bonnie Webber, Luke Zettlemoyer, and the members of ILCC Edinburgh University, the Microsoft Research Redmond NLP group, the Stanford NLP group, and the UW NLP and Linguistics group. We thank Reviewer 2 for useful feedback. The authors would also like to thank the Universal Dependencies community for the treebanks and documentation. This research is supported by a Google PhD Fellowship to the first author. We acknowledge the financial support of the European Research Council (Lapata; award number 681760).

## References

Lasha Abzianidze, Johannes Bjerva, Kilian Evang, Hessel Haagsma, Rik van Noord, Pierre Ludmann, Duc-Duy Nguyen, and Johan Bos. 2017. The Parallel Meaning Bank: Towards a Multilingual Corpus of Translations Annotated with Compositional Meaning Representations. In *Proceedings of the*

*European Chapter of the Association for Computational Linguistics*. Association for Computational Linguistics, Valencia, Spain, pages 242–247.

Alan Akbik, laura chiticariu, Marina Danilevsky, Yunyao Li, Shivakumar Vaithyanathan, and Huaiyu Zhu. 2015. Generating High Quality Proposition Banks for Multilingual Semantic Role Labeling. In *Proceedings of the Association for Computational Linguistics and the International Joint Conference on Natural Language Processing*. Association for Computational Linguistics, Beijing, China, pages 397–407.

Rami Al-Rfou, Bryan Perozzi, and Steven Skiena. 2013. Polyglot: Distributed Word Representations for Multilingual NLP. In *Proceedings of the Computational Natural Language Learning*. Sofia, Bulgaria, pages 183–192.

Yoav Artzi. 2013. Cornell SPF: Cornell Semantic Parsing Framework. *arXiv:1311.3011 [cs.CL]*.

Yoav Artzi, Kenton Lee, and Luke Zettlemoyer. 2015. Broad-coverage CCG Semantic Parsing with AMR. In *Proceedings of the Empirical Methods on Natural Language Processing*. pages 1699–1710.

Jason Baldridge and Geert-Jan Kruijff. 2002. Coupling CCG and Hybrid Logic Dependency Semantics. In *Proceedings of the Association for Computational Linguistics*. pages 319–326.

Laura Banarescu, Claire Bonial, Shu Cai, Madalina Georgescu, Kira Griffitt, Ulf Hermjakob, Kevin Knight, Philipp Koehn, Martha Palmer, and Nathan Schneider. 2013. Abstract Meaning Representation for Sembanking. In *Linguistic Annotation Workshop and Interoperability with Discourse*. Sofia, Bulgaria, pages 178–186.

Hannah Bast and Elmar Haussmann. 2015. More Accurate Question Answering on Freebase. In *Proceedings of ACM International Conference on Information and Knowledge Management*. pages 1431–1440.

Paul Bédaride and Claire Gardent. 2011. Deep Semantics for Dependency Structures. In *Proceedings of Conference on Intelligent Text Processing and Computational Linguistics*. pages 277–288.

Islam Beltagy. 2016. *Natural Language Semantics Using Probabilistic Logic*. Ph.D. thesis, Department of Computer Science, The University of Texas at Austin.

Emily M. Bender, Dan Flickinger, Stephan Oepen, Woodley Packard, and Ann Copestake. 2015. Layers of Interpretation: On Grammar and Compositionality. In *Proceedings of the International Conference on Computational Semantics*. Association for Computational Linguistics, London, UK, pages 239–249.

Jonathan Berant, Andrew Chou, Roy Frostig, and Percy Liang. 2013. Semantic Parsing on Freebase from Question-Answer Pairs. In *Proceedings of the Empirical Methods on Natural Language Processing*. pages 1533–1544.

Jonathan Berant and Percy Liang. 2014. Semantic Parsing via Paraphrasing. In *Proceedings of the Association for Computational Linguistics*. pages 1415–1425.

Jonathan Berant and Percy Liang. 2015. Imitation Learning of Agenda-Based Semantic Parsers. *Transactions of the Association for Computational Linguistics* 3:545–558.

Anders Bjrkelund, Love Hafdell, and Pierre Nugues. 2009. Multilingual Semantic Role Labeling. In *Proceedings of Computational Natural Language Learning (CoNLL 2009): Shared Task*. Association for Computational Linguistics, Boulder, Colorado, pages 43–48.

Antoine Bordes, Nicolas Usunier, Sumit Chopra, and Jason Weston. 2015. Large-scale simple question answering with memory networks. CoRR abs/1506.02075.

Johan Bos, Stephen Clark, Mark Steedman, James R. Curran, and Julia Hockenmaier. 2004. Wide-Coverage Semantic Representations from a CCG Parser. In *Proceedings of the Conference on Computational Linguistics*. pages 1240–1246.

Philipp Cimiano, Vanessa Lopez, Christina Unger, Elena Cabrio, Axel-Cyrille Ngonga Ngomo, and Sebastian Walter. 2013. Multilingual question answering over linked data (QALD-3): Lab overview. In *Information Access Evaluation. Multilinguality, Multimodality, and Visualization*. Springer, Valencia, Spain, volume 8138.

Michael Collins. 2002. Discriminative Training Methods for Hidden Markov Models: Theory and Experiments with Perceptron Algorithms. In *Proceedings of the Empirical Methods on Natural Language Processing*. pages 1–8.

Ronan Collobert, Jason Weston, Leon Bottou, Michael Karlen, Koray Kavukcuoglu, and Pavel Kuks. 2011. Natural language processing (almost) from scratch. *Journal of Machine Learning Research* 12:2493–2537.

Ann Copestake, Dan Flickinger, Carl Pollard, and Ivan A. Sag. 2005. Minimal Recursion Semantics: An Introduction. *Research on Language and Computation* 3(2-3):281–332.

Dick Crouch and Tracy Holloway King. 2006. Semantics via f-structure rewriting. In *Proceedings of the LFG'06 Conference*. CSLI Publications, page 145.

Mary Dalrymple, John Lamping, Fernando C. N. Pereira, and Vijay A. Saraswat. 1995. Linear Logic for Meaning Assembly. In *Proceedings of Computational Logic for Natural Language Processing*.

Kilian Evang and Johan Bos. 2016. Cross-lingual Learning of an Open-domain Semantic Parser. In *Proceedings of the Conference on Computational Linguistics*. The COLING 2016 Organizing Committee, Osaka, Japan, pages 579–588.

Tobias Falke, Gabriel Stanovsky, Iryna Gurevych, and Ido Dagan. 2016. Porting an Open Information Extraction System from English to German. In *Proceedings of the Empirical Methods in Natural Language Processing*. Association for Computational Linguistics, Austin, Texas, pages 892–898.

Jeffrey Flanigan, Sam Thomson, Jaime Carbonell, Chris Dyer, and Noah A. Smith. 2014. A Discriminative Graph-Based Parser for the Abstract Meaning Representation. In *Proceedings of the Association for Computational Linguistics*. pages 1426–1436.

Claire Gardent and Laura Kallmeyer. 2003. Semantic Construction in Feature-based TAG. In *Proceedings of European Chapter of the Association for Computational Linguistics*. pages 123–130.

Matt Gardner and Jayant Krishnamurthy. 2017. Open-Vocabulary Semantic Parsing with both Distributional Statistics and Formal Knowledge. In *Proceedings of Association for the Advancement of Artificial Intelligence*.

Jonas Groschwitz, Alexander Koller, and Christoph Teichmann. 2015. Graph parsing with s-graph grammars. In *Proceedings of the Association for Computational Linguistics*. pages 1481–1490.

Carolin Haas and Stefan Riezler. 2016. A Corpus and Semantic Parser for Multilingual Natural Language Querying of OpenStreetMap. In *Proceedings of the North American Chapter of the Association for Computational Linguistics: Human Language Technologies*. Association for Computational Linguistics, San Diego, California, pages 740–750.

Michael Hahn and Detmar Meurers. 2011. On deriving semantic representations from dependencies: A practical approach for evaluating meaning in learner corpora. In *Proceedings of the Int. Conference on Dependency Linguistics (Depling 2011)*. Barcelona, pages 94–103.

Jan Haji, Massimiliano Ciaramita, Richard Johansson, Daisuke Kawahara, Maria Antnia Mart, Llus Mrquez, Adam Meyers, Joakim Nivre, Sebastian Pad, Jan tpnek, and others. 2009. The CoNLL-2009 shared task: Syntactic and semantic dependencies in multiple languages. In *Proceedings of the Computational Natural Language Learning: Shared Task*. Association for Computational Linguistics, pages 1–18.

Kejun Huang, Matt Gardner, Evangelos Papalexakis, Christos Faloutsos, Nikos Sidiropoulos, Tom Mitchell, Partha P. Talukdar, and Xiao Fu. 2015. Translation Invariant Word Embeddings. In *Proceedings of the Empirical Methods in Natural Language Processing*. Lisbon, Portugal, pages 1084–1088.

Max Jakob, Markéta Lopatková, and Valia Kordoni. 2010. Mapping between Dependency Structures and Compositional Semantic Representations. In *Proceedings of the Fifth International Conference on Language Resources and Evaluation*.

Zhanming Jie and Wei Lu. 2014. Multilingual Semantic Parsing : Parsing Multiple Languages into Semantic Representations. In *Proceedings of the Conference on Computational Linguistics*. Dublin City University and Association for Computational Linguistics, Dublin, Ireland, pages 1291–1301.

Bevan Keeley Jones, Mark Johnson, and Sharon Goldwater. 2012. Semantic Parsing with Bayesian Tree Transducers. In *Proceedings of the Association for Computational Linguistics*. Association for Computational Linguistics, Stroudsburg, PA, USA, pages 488–496.

Laura Kallmeyer and Aravind Joshi. 2003. Factoring predicate argument and scope semantics: Underspecified semantics with LTAG. *Research on Language and Computation* 1(1-2):3–58.

Ronald M Kaplan and Joan Bresnan. 1982. Lexical-functional grammar: A formal system for grammatical representation. *Formal Issues in Lexical-Functional Grammar* pages 29–130.

Eliyahu Kiperwasser and Yoav Goldberg. 2016. Simple and Accurate Dependency Parsing Using Bidirectional LSTM Feature Representations. *Transactions of the Association for Computational Linguistics* 4:313–327.

Jayant Krishnamurthy and Tom M. Mitchell. 2015. Learning a Compositional Semantics for Freebase with an Open Predicate Vocabulary. *Transactions of the Association for Computational Linguistics* 3:257–270.

Tom Kwiatkowksi, Luke Zettlemoyer, Sharon Goldwater, and Mark Steedman. 2010. Inducing Probabilistic CCG Grammars from Logical Form with Higher-Order Unification. In *Proceedings of the Empirical Methods on Natural Language Processing*. pages 1223–1233.

Tom Kwiatkowski, Eunsol Choi, Yoav Artzi, and Luke Zettlemoyer. 2013. Scaling Semantic Parsers with On-the-Fly Ontology Matching. In *Proceedings of the Empirical Methods on Natural Language Processing*. pages 1545–1556.

Roger Levy and Galen Andrew. 2006. Tregex and tsurgeon: tools for querying and manipulating tree data structures. In *Proceedings of LREC*. pages 2231–2234.

Fei Liu, Jeffrey Flanigan, Sam Thomson, Norman Sadeh, and Noah A. Smith. 2015. Toward Abstractive Summarization Using Semantic Representations. In *Proceedings of North American Chapter of the Association for Computational Linguistics*. pages 1077–1086.

Mitchell P. Marcus, Mary Ann Marcinkiewicz, and Beatrice Santorini. 1993. Building a large annotated corpus of English: The Penn Treebank. *Computational linguistics* 19(2):313–330.

Richard Montague. 1973. The Proper Treatment of Quantification in Ordinary English. In K.J.J. Hintikka, J.M.E. Moravcsik, and P. Suppes, editors, *Approaches to Natural Language*, Springer Netherlands, volume 49 of *Synthese Library*, pages 221–242.

Rebecca Nesson and Stuart M. Shieber. 2006. Simpler TAG Semantics Through Synchronization. In *Proceedings of the 11th Conference on Formal Grammar*. Center for the Study of Language and Information, Malaga, Spain, pages 129–142.

Joakim Nivre, Marie-Catherine de Marneffe, Filip Ginter, Yoav Goldberg, Jan Hajic, Christopher D. Manning, Ryan McDonald, Slav Petrov, Sampo Pyysalo, Natalia Silveira, Reut Tsarfaty, and Daniel Zeman. 2016. Universal Dependencies v1: A Multilingual Treebank Collection. In *Proceedings of the Tenth International Conference on Language Resources and Evaluation*. European Language Resources Association (ELRA), Paris, France.

Joakim Nivre et al. 2016. Universal dependencies 1.3. LINDAT/CLARIN digital library at the Institute of Formal and Applied Linguistics, Charles University in Prague.

Martha Palmer, Daniel Gildea, and Paul Kingsbury. 2005. The proposition bank: An annotated corpus of semantic roles. *Computational linguistics* 31(1):71–106.

Martha Palmer, Daniel Gildea, and Nianwen Xue. 2010. Semantic role labeling. *Synthesis Lectures on Human Language Technologies* 3(1):1–103.

Panupong Pasupat and Percy Liang. 2015. Compositional Semantic Parsing on Semi-Structured Tables. In *Proceedings of the Association for Computational Linguistics*. pages 1470–1480.

Jeffrey Pennington, Richard Socher, and Christopher Manning. 2014. Glove: Global Vectors for Word Representation. In *Proceedings of the Empirical Methods in Natural Language Processing*. Association for Computational Linguistics, Doha, Qatar, pages 1532–1543.

Barbara Plank, Anders Søgaard, and Yoav Goldberg. 2016. Multilingual Part-of-Speech Tagging with Bidirectional Long Short-Term Memory Models and Auxiliary Loss. In *Proceedings of the Annual Meeting of the Association for Computational Linguistics*. Berlin, Germany, pages 412–418.

Siva Reddy, Mirella Lapata, and Mark Steedman. 2014. Large-scale Semantic Parsing without Question-Answer Pairs. *Transactions of the Association for Computational Linguistics* 2:377–392.

Siva Reddy, Oscar Täckström, Michael Collins, Tom Kwiatkowski, Dipanjan Das, Mark Steedman, and Mirella Lapata. 2016. Transforming Dependency Structures to Logical Forms for Semantic Parsing. *Transactions of the Association for Computational Linguistics* 4:127–140.

Michael Roth and Kristian Woodsend. 2014. Composition of Word Representations Improves Semantic Role Labelling. In *Proceedings of the Empirical Methods in Natural Language Processing (EMNLP)*. Association for Computational Linguistics, Doha, Qatar, pages 407–413.

Mrinmaya Sachan and Eric Xing. 2016. Machine Comprehension using Rich Semantic Representations. In *Proceedings of the Association for Computational Linguistics*. Association for Computational Linguistics, Berlin, Germany, pages 486–492.

Sebastian Schuster and Christopher D. Manning. 2016. Enhanced English Universal Dependencies: An Improved Representation for Natural Language Understanding Tasks. In *Proceedings of the Tenth International Conference on Language Resources and Evaluation*. European Language Resources Association (ELRA), Paris, France.

Kiril Simov and Petya Osenova. 2011. Towards Minimal Recursion Semantics over Bulgarian Dependency Parsing. In *Proceedings of the International Conference Recent Advances in Natural Language Processing 2011*. RANLP 2011 Organising Committee, Hissar, Bulgaria, pages 471–478.

Kathrin Spreyer and Anette Frank. 2005. Projecting RMRS from TIGER Dependencies. In *Proceedings of the HPSG 2005 Conference*. CSLI Publications.

Yu Su, Huan Sun, Brian Sadler, Mudhakar Srivatsa, Izzeddin Gur, Zenghui Yan, and Xifeng Yan. 2016. On Generating Characteristic-rich Question Sets for QA Evaluation. In *Proceedings of the Empirical Methods in Natural Language Processing*. Austin, Texas, pages 562–572.

Ferhan Türe and Oliver Jojic. 2016. Simple and Effective Question Answering with Recurrent Neural Networks. *CoRR* abs/1606.05029.

Lucy Vanderwende, Arul Menezes, and Chris Quirk. 2015. An AMR parser for English, French, German, Spanish and Japanese and a new AMR-annotated corpus. In *Proceedings of the North American Chapter of the Association for Computational Linguistics: Demonstrations*. Association for Computational Linguistics, Denver, Colorado, pages 26–30.

Aaron Steven White, Drew Reisinger, Keisuke Sakaguchi, Tim Vieira, Sheng Zhang, Rachel Rudinger, Kyle Rawlins, and Benjamin Van Durme. 2016. Universal Decompositional Semantics on Universal Dependencies. In *Proceedings of the Empirical Methods in Natural Language Processing*. Association for Computational Linguistics, Austin, Texas, pages 1713–1723.

Kun Xu, Siva Reddy, Yansong Feng, Songfang Huang, and Dongyan Zhao. 2016. Question Answering on Freebase via Relation Extraction and Textual Evidence. In *Proceedings of the Association for Computational Linguistics*. Association for Computational Linguistics, Berlin, Germany, pages 2326–2336.

Xuchen Yao. 2015. Lean Question Answering over Freebase from Scratch. In *Proceedings of North American Chapter of the Association for Computational Linguistics*. pages 66–70.

Xuchen Yao and Benjamin Van Durme. 2014. Information Extraction over Structured Data: Question Answering with Freebase. In *Proceedings of the Association for Computational Linguistics*. pages 956–966.

Semih Yavuz, Izzeddin Gur, Yu Su, Mudhakar Srivatsa, and Xifeng Yan. 2016. Improving Semantic Parsing via Answer Type Inference. In *Proceedings of the Empirical Methods in Natural Language Processing*. Association for Computational Linguistics, Austin, Texas, pages 149–159.

Wen-tau Yih, Ming-Wei Chang, Xiaodong He, and Jianfeng Gao. 2015. Semantic Parsing via Staged Query Graph Generation: Question Answering with Knowledge Base. In *Proceedings of the Association for Computational Linguistics*. pages 1321–1331.

John M. Zelle and Raymond J. Mooney. 1996. Learning to Parse Database Queries Using Inductive Logic Programming. In *Proceedings of Association for the Advancement of Artificial Intelligence*. pages 1050–1055.

Luke S. Zettlemoyer and Michael Collins. 2005. Learning to Map Sentences to Logical Form: Structured Classification with Probabilistic Categorial Grammars. In *Proceedings of Uncertainty in Artificial Intelligence*. pages 658–666.

# Universal Semantic Parsing: Supplementary Material

Siva Reddy[†]  Oscar Täckström[‡]  Slav Petrov[‡]  Mark Steedman[††]  Mirella Lapata[††]
[†]Stanford University
[‡] Google Inc.
[††]University of Edinburgh
sivar@stanford.edu, {oscart, slav}@google.com, {steedman, mlap}@inf.ed.ac.uk

## Abstract

This supplementary material to the main paper, provides an outline of how quantification can be incorporated in the UDEP-LAMBDA framework.

## 1 Universal Quantification

Consider the sentence *Everybody wants to buy a house*,[1] whose dependency tree in the Universal Dependencies (UD) formalism is shown in Figure 1(a). This sentence has two possible readings: either (1) every person wants to buy a different house; or (2) every person wants to buy the same house. The two interpretations correspond to the following logical forms:

(1) $\forall x.\, \text{person}(x_a) \rightarrow$
    $[\exists zyw.\, \text{wants}(z_e) \wedge \text{arg}_1(z_e, x_a) \wedge \text{buy}(y_e) \wedge \text{xcomp}(z_e, y_e) \wedge$
    $\text{house}(w_a) \wedge \text{arg}_1(z_e, x_a) \wedge \text{arg}_2(z_e, w_a)]$ ;

(2) $\exists w.\, \text{house}(w_a) \wedge (\forall x.\, \text{person}(x_a) \rightarrow$
    $[\exists zy.\, \text{wants}(z_e) \wedge \text{arg}_1(z_e, x_a) \wedge \text{buy}(y_e) \wedge \text{xcomp}(z_e, y_e) \wedge$
    $\text{arg}_1(z_e, x_a) \wedge \text{arg}_2(z_e, w_a)])$ .

In (1), the existential variable *w* is in the scope of the universal variable *x* (i.e. the *house* is dependent on the *person*). This reading is commonly referred to as the *surface reading*. Conversely, in (2) the universal variable *x* is in the scope of the existential variable *w* (i.e. the *house* is independent of the *person*). This reading is also called *inverse reading*. Our goal is to obtain the surface reading logical form in (1) with UDEPLAMBDA. We do not aim to obtain the inverse reading, although this is possible with the use of Skolemization (Steedman, 2012).

In UDEPLAMBDA, lambda expressions for words, phrases and sentences are all of the form $\lambda x.\,\ldots$. But from (1), it is clear that we need to express variables bound by quantifiers, e.g. $\forall x$, while still providing access to *x* for composition. This demands a change in the type system since the

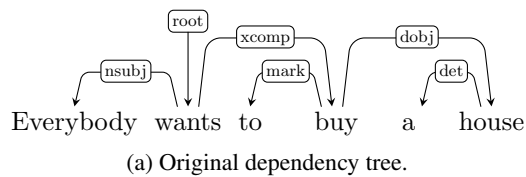

(a) Original dependency tree.

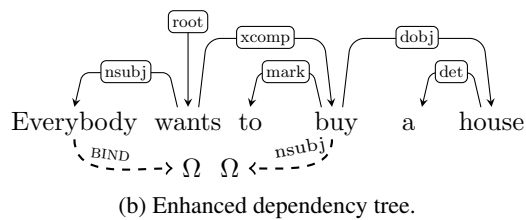

(b) Enhanced dependency tree.

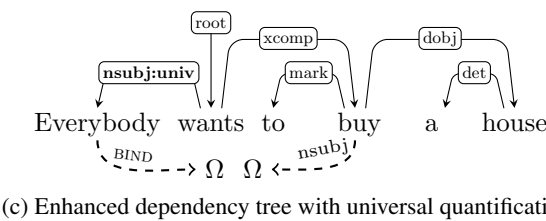

(c) Enhanced dependency tree with universal quantification.

Figure 1: The dependency tree for *Everybody wants to buy a house* and its enhanced variants.

same variable cannot be lambda bound and quantifier bound—that is we cannot have formulas of the form $\lambda x \ldots \forall x \ldots$. In this material, we first derive the logical form for the example sentence using the type system from our main paper (Section 1.1) and show that it fails to handle universal quantification. We then modify the type system slightly to allow derivation of the desired surface reading logical form (Section 1.2). This modified type system is a strict generalization of the original type system.[2] Fancellu et al. (2017) present an elaborate discussion on the modified type system, and how it can handle negation scope and its interaction with universal quantifiers.

---

[1]Example borrowed from Schuster and Manning (2016).

[2]Note that this treatment has yet to be added to our implementation, which can be found at https://github.com/sivareddyg/udeplambda.

## 1.1 With Original Type System

We will first attempt to derive the logical form in (1) using the default type system of UDEPLAMBDA. Figure 1(b) shows the enhanced dependency tree for the sentence, where BIND has been introduced to connect the implied `nsubj` of *buy* (BIND is explained in the main paper in Section 3.2). The s-expression corresponding to the enhanced tree is:

(nsubj (xcomp wants (mark
        (nsubj (dobj buy (det house a)) Ω) to))
    (BIND everybody Ω)) .

With the following substitution entries,

*wants*, *buy* ∈ EVENT;
*everybody*, *house* ∈ ENTITY;
*a*, *to* ∈ FUNCTIONAL;
$\Omega = \lambda x. \text{EQ}(x, \omega)$;
$\texttt{nsubj} = \lambda fgx. \exists y. f(x) \wedge g(y) \wedge \arg_1(x_e, y_a)$;
$\texttt{dobj} = \lambda fgx. \exists y. f(x) \wedge g(y) \wedge \arg_2(x_e, y_a)$;
$\texttt{xcomp} = \lambda fgx. \exists y. f(x) \wedge g(y) \wedge \text{xcomp}(x_e, y_a)$;
`mark` ∈ HEAD;
BIND ∈ MERGE,

the lambda expression after composition becomes:

$\lambda z. \exists xywv. \text{wants}(z_e) \wedge \text{everybody}(x_a) \wedge \arg_1(z_e, x_a)$
    $\wedge \text{EQ}(x, \omega) \wedge \text{buy}(y_e) \wedge \text{xcomp}(z_e, y_e) \wedge \arg_1(y_e, v_a)$
    $\wedge \text{EQ}(v, \omega) \wedge \arg_1(x_e, y_a) \wedge \text{house}(w_a) \wedge \arg_2(y_e, w_a)$.

This expression encodes the fact that *x* and *v* are in unification, and can thus be further simplified to:

(3) $\lambda z. \exists xyw. \text{wants}(z_e) \wedge \text{everybody}(x_a) \wedge \arg_1(z_e, x_a)$
    $\wedge \text{buy}(y_e) \wedge \text{xcomp}(z_e, y_e) \wedge \arg_1(y_e, x_a)$
    $\wedge \arg_1(x_e, y_a) \wedge \text{house}(w_a) \wedge \arg_2(y_e, w_a)$.

However, the logical form (3) differs from the desired form (1). As noted above, UDEPLAMBDA with its default type, where each s-expression must have the type η = **Ind** × **Event** → **Bool**, cannot handle quantifier scoping.

## 1.2 With Higher-order Type System

Following Champollion (2010), we make a slight modification to the type system. Instead of using expressions of the form $\lambda x. \ldots$ for words, we use either $\lambda f. \exists x. \ldots$ or $\lambda f. \forall x. \ldots$, where *f* has type η. As argued by Champollion, this higher-order form makes quantification and negation handling sound and simpler in Neo-Davidsonian event semantics. Following this change, we assign the following lambda expressions to the words in our example sentence:

*everybody* $= \lambda f. \forall x. \text{person}(x) \rightarrow f(x)$;
*wants* $= \lambda f. \exists x. \text{wants}(x_e) \wedge f(x)$;
*to* $= \lambda f. \text{TRUE}$;
*buy* $= \lambda f. \exists x. \text{buy}(x_e) \wedge f(x)$;
*a* $= \lambda f. \text{TRUE}$;
*house* $= \lambda f. \exists x. \text{house}(x_a) \wedge f(x)$;
$\Omega = \lambda f. f(\omega)$.

Here *everybody* is assigned universal quantifier semantics. Since the UD representation does not distinguish quantifiers, we need to rely on a small (language-specific) lexicon to identify these. To encode quantification scope, we enhance the label `nsubj` to `nsubj:univ`, which indicates that the subject argument of *wants* contains a universal quantifier, as shown in Figure 1(c).

This change of semantic type for words and s-expressions forces us to also modify the semantic type of dependency labels, in order to obey the single-type constraint of DEPLAMBDA (Reddy et al., 2016). Thus, dependency labels will now take the form $\lambda PQf. \ldots$, where *P* is the parent expression, *Q* is the child expression, and the *return expression* is of the form $\lambda f. \ldots$. Following this change, we assign the following lambda expressions to dependency labels:

$\texttt{nsubj:univ} = \lambda PQf. Q(\lambda y. P(\lambda x. f(x) \wedge \arg_1(x_e, y_a)))$;
$\texttt{nsubj} = \lambda PQf. P(\lambda x. f(x) \wedge Q(\lambda y. \arg_1(x_e, y_a)))$;
$\texttt{dobj} = \lambda PQf. P(\lambda x. f(x) \wedge Q(\lambda y. \arg_2(x_e, y_a)))$;
$\texttt{xcomp} = \lambda PQf. P(\lambda x. f(x) \wedge Q(\lambda y. \text{xcomp}(x_e, y_a)))$;
$\texttt{det}, \texttt{mark} = \lambda PQf. P(f)$;
$\text{BIND} = \lambda PQf. P(\lambda x. f(x) \wedge Q(\lambda y. \text{EQ}(y, x)))$.

Notice that the lambda expression of `nsubj:univ` differs from `nsubj`. In the former, the lambda variables inside *Q* have wider scope over the variables in *P* (i.e. the universal quantifier variable of *everybody* has scope over the event variable of *wants*) contrary to the latter.

The new s-expression for Figure 1(c) is

(nsubj:univ (xcomp wants (mark
        (nsubj (dobj buy (det house a)) Ω) to))
    (BIND everybody Ω)) .

Substituting with the modified expressions, and performing composition and simplification leads to the expression:

(6) $\lambda f. \forall x . \text{person}(x_a) \rightarrow$
    $[\exists zyw. f(z) \wedge \text{wants}(z_e) \wedge \arg_1(z_e, x_a) \wedge \text{buy}(y_e)$
    $\wedge \text{xcomp}(z_e, y_e) \wedge \text{house}(w_a)$
    $\wedge \arg_1(z_e, x_a) \wedge \arg_2(z_e, w_a)]$.

This expression is identical to (1) except for the outermost term $\lambda f$. By applying (6) to $\lambda x. \text{TRUE}$, we obtain (1), which completes the treatment of universal quantification in UDEPLAMBDA.

## References

Lucas Champollion. 2010. Quantification and negation in event semantics. *Baltic International Yearbook of Cognition, Logic and Communication* 6(1):3.

Federico Fancellu, Siva Reddy, Adam Lopez, and Bonnie Webber. 2017. Universal Dependencies to Logical Forms with Negation Scope. *arXiv Preprint* .

Siva Reddy, Oscar Täckström, Michael Collins, Tom Kwiatkowski, Dipanjan Das, Mark Steedman, and Mirella Lapata. 2016. Transforming Dependency Structures to Logical Forms for Semantic Parsing. *Transactions of the Association for Computational Linguistics* 4:127–140.

Sebastian Schuster and Christopher D. Manning. 2016. Enhanced English Universal Dependencies: An Improved Representation for Natural Language Understanding Tasks. In *Proceedings of the Tenth International Conference on Language Resources and Evaluation*. European Language Resources Association (ELRA), Paris, France.

Mark Steedman. 2012. *Taking Scope - The Natural Semantics of Quantifiers*. MIT Press.